\documentclass[conference]{IEEEtran}
\IEEEoverridecommandlockouts
\usepackage{cite}
\usepackage{amsmath,amssymb,amsfonts}
\usepackage{algorithmic}
\usepackage{graphicx}
\usepackage{textcomp}
\usepackage{xcolor}
\usepackage{array}

\usepackage{enumerate}
\usepackage{url}
\usepackage{multirow}
\usepackage{booktabs}

\usepackage{todonotes}

\usepackage{xurl}

\def\BibTeX{{\rm B\kern-.05em{\sc i\kern-.025em b}\kern-.08em
    t\kern-.1667em\lower.7ex\hbox{E}\kern-.125emX}}

\usepackage{fancyhdr}
\begin{document}

\title{An Ethical Framework for Guiding the Development of 
Affectively-Aware Artificial Intelligence
\thanks{This research is supported in part by the National Research Foundation, Singapore under its AI Singapore Program (AISG Award No: AISG2-RP-2020-016), and a Singapore Ministry of Education Academic Research Fund Tier 1 grant to DCO.}
} 

\author{
\IEEEauthorblockN{
Desmond~C.~Ong
}
\IEEEauthorblockA{
\textit{Department of Information Systems and Analytics, National University of Singapore} \\
\textit{Institute of High Performance Computing, Agency for Science, Technology and Research, Singapore} \\
dco@comp.nus.edu.sg}
}

\maketitle
\thispagestyle{fancy}

\begin{abstract}

The recent rapid advancements in artificial intelligence research and deployment have sparked more discussion about the potential ramifications of socially- and emotionally-intelligent AI. The question is not if research can produce such affectively-aware AI, but when it will. What will it mean for society when machines---and the corporations and governments they serve---can ``read" people’s minds and emotions? What should developers and operators of such AI do, and what should they not do? The goal of this article is to pre-empt some of the potential implications of these developments, and propose a set of guidelines for evaluating the (moral and) ethical consequences of affectively-aware AI, in order to guide researchers, industry professionals, and policy-makers. We propose a multi-stakeholder analysis framework that separates the ethical responsibilities of AI Developers vis-\`a-vis the entities that deploy such AI---which we term Operators. Our analysis produces two pillars that clarify the responsibilities of each of these stakeholders: \textit{Provable Beneficence}, which rests on proving the effectiveness of the AI, and \textit{Responsible Stewardship}, which governs responsible collection, use, and storage of data and the decisions made from such data.
We end with recommendations for researchers, developers, operators, as well as regulators and law-makers. 

\end{abstract}


\section{Introduction}

At the February 2020 meeting of the Association for the Advancement of Artificial Intelligence (AAAI), the premier society for AI research, two invited plenary speakers mentioned Affective Computing in their talks, for completely opposite reasons. The first speaker cited Affective Computing as an example of a world-destroying AI application, as it could empower authoritarian regimes to monitor the inner lives of their citizens. Thirty-six hours later, a second speaker hailed it as a crucial component to building provably safe AI and preventing an AI-led apocalypse\footnote{Henry Kautz cited the threat of AI that can infer moods, beliefs, and intentions from data, which could be used by state actors to target and suppress dissidents. Stuart Russell proposed that inferring emotions is critical to solving the value-alignment problem (aligning AI utility functions with human utility functions), which in turn is necessary for provably-safe AI.}. Neither studies affective computing; both are respected AI researchers whose opinions about affective computing could not be further apart.

Any technology can be applied productively, or in questionable ways. In the case of affective computing, conversations about mis-uses are happening more frequently. 
For example, in the interests of more efficiently screening large quantities of job applicants, companies around the world are utilizing emotion recognition in AI for automated candidate assessment \cite{cha2020smile, metz2020there, fernandez2020ai}. This has faced backlash \cite{chen2020emotion, harwell2019face} and has even generated legislation \cite{ai-video-act-2019}. Other controversial examples include emotion detection for employee and student monitoring \cite{dzieza2020emotion, andrejevic2020facial, li2019brain}. Psychologists have also questioned whether current emotion recognition models are scientifically validated enough to afford the inferences that companies are drawing \cite{barrett2019emotional}. This has led some AI researchers to start calling for an outright ban on deploying emotion recognition technologies in ``decisions that impact people's lives and access to opportunities" \cite{crawford2019ai}.
At the last \textit{Affective Computing and Intelligent Interaction} meeting in September 2019, we---the community of affective computing researchers---convened an inaugural townhall to discuss problematic applications of affective computing technology, and we decided that the conference theme for 2021 would be \textit{Ethical Affective Computing}. Indeed, many of us feel that we, being the most familiar with the capabilities and limitations of such technology, as well as having done the research that enabled these applications, have a professional responsibility to address these ethical issues head-on.

The goal of this article is to start a conversation on systematically addressing these ethical issues. 
Building upon past discussions in affective computing \cite{cowie2012good, mcstay2018emotional, cowie2015ethical, sedenberg2017smile, stark2021ethics} and current trends in AI ethics more broadly \cite{jobin2019global}, we outline an ethical framework for examining the impact of applications of affectively-aware AI. Ethics should not just prescribe what we should \emph{not} do with such technology, but also illuminate what we \emph{should} do---and how we should do it. This framework aims to serve as a set of guidelines that relevant stakeholders, including researchers, industry professionals, companies, policy-makers, and regulators, can employ to analyze the associated risks, which is crucial to a human-centered approach to developing affectively-aware AI technologies.

\subsection{Scope}
In this article we specifically focus on issues particular to affectively-aware AI, and avoid discussion of the ethics of AI more generally: We note that of the 84 documents identified by \cite{jobin2019global}'s extensive review of AI ethics statements, only a minority (11 documents, or 13\%) \cite{crawford2019ai, whittaker2018ai, ead2016, inria2014ethique, australia2019ai, france2017how, france2018for, dutch2017human, dutch2018dutch, comest2017report, campolo2017} mentioned emotion recognition AI, and only 4 \cite{crawford2019ai, whittaker2018ai, ead2016, inria2014ethique} and \cite{mcstay2019guidelines} discuss it in detail. We use the term Affectively-Aware AI 
to specifically refer to AI that can recognize emotional expressions in people (from facial expressions, body language, language, physiology, and so forth), and perhaps can reason about emotions in context or with cultural awareness. 
Although the terms ``Affective AI" or ``Emotional AI" are commonly used by affective computing researchers and companies, we specifically choose not to use them to sidestep any philosophical discussion of AI that \emph{possesses} emotions \cite{beavers2017moral}. We also do not discuss ethical issues with AI expressing emotions (e.g., deception \cite{cowie2015ethical}).
The consequences we discuss rest only on the presumed ability of AI to recognize emotions---Indeed, emotion-recognizing AI make up the vast majority of commercial offerings that already exist today, and that are generating public concern. 

We first begin by briefly summarizing the capabilities and limitations of the technology today, as overblown claims by companies have led to a general misunderstanding about the limits of the technology \cite{chen2020emotion}. Next, we outline our proposed ethical framework, which analyzes the different stakeholders relevant to affectively-aware AI. In particular, we distinguish the ethical responsibilities held by the \textbf{AI Developer} with the entities that deploy such AI (which we term \textbf{Operators}), in order to clarify the division of responsibilities, and avoid either party absolving responsibility. We describe the two broad ethical pillars in our framework, and the implications that they have on development and deployment of such AI.


\subsection{Why the need for Ethical Guidelines}
Ethical guidelines are necessary to help guide what professionals \emph{should} do with the technology they create, in addition to what they \emph{should not} do. Although ethical guidelines by themselves carry no ``hard" power---unlike laws and regulations passed by government authorities---they do serve an important role in clarifying social and professional norms. Articulating a clear and unambiguous ethical code provides clarification for activities the field deems acceptable and those it does not, and will help to guide both junior and seasoned researchers. Second, ethical guidelines could inspire individual entities (academic departments, journals, conferences, grant funding agencies) to implement policies that encourage compliance (e.g., a focus on teaching ethics in training curricula, or requiring ethical analysis to supplement paper submissions). Finally, putting forward a researcher-written framework may help regulators align efforts to craft complementary laws. 

Many professions and professional organizations have their own ethical codes by which they expect practicing members to adhere to. Perhaps the oldest and most famous is the Hippocratic Oath, and modern versions of this oath exist in many medical schools today. The Association for Computing Machinery (ACM) and the Institute of Electrical and Electronics Engineers (IEEE), the two largest engineering professional bodies, similarly have Codes of Ethics, which focus broadly on principles such as honesty, fairness, and respect for privacy. 

More specific to Affective Computing, the IEEE Global Initiative on Ethics of Autonomous and Intelligent Systems---a committee of respected IEEE engineers---published ``Ethically Aligned Design" (EAD), a document focusing broadly on AI ethical issues, and which has a chapter dedicated to Affective Computing \cite{ead2016}. This chapter outlines recommendations for specific application areas: (i) affective systems that are deployed across cultures; (ii) affective systems with which users may develop relationships; (iii) affective systems which manipulate, ``nudge" or deceive people; (iv) affective systems which may impact human autonomy when deployed in organizations or in societies; and (v) affective systems which display synthetic emotions. 

In order to complement EAD's approach of focusing on issues specific to individual applications, the present paper first outlines a more general, multi-stakeholder framework, elaborating on the responsibilities of both the AI developers and the entities that deploy such AI. 
Indeed, our proposed pillars encapsulates all the principles elaborated on in EAD\footnote{Our first pillar includes EAD's General Principles of \emph{Human Rights}, \emph{Well-Being}, \emph{Effectiveness}, \emph{Transparency}, and \emph{Accountability}; while our second includes the remainder: \emph{Data Agency}, \emph{Awareness of Misuse}, and \emph{Competence}.}, 
and add several insights and concrete recommendations from reasoning through the two main stakeholders.
In doing so, we hope that this document will serve as an elaboration of widely-accepted ethical principles, that is also actionable by researchers, industry professionals and policy-makers.

\subsection{Summarizing recent advancements in affectively-aware AI}

Trying to characterize the state of a rapidly-moving field, and so briefly, is a Sisyphean task. However, we feel that it is important, especially for readers outside affective computing, to know what are the capabilities and the limitations of this technology today. 
There are many scientific/psychological theories of emotion \cite{gross2011emotion}, such as basic emotion theories \cite{ekman1999basic}, appraisal theories \cite{moors2013appraisal}, and constructivist theories \cite{mesquita2010contextualized}. Although a full discussion of these theories is beyond our scope, we invite the interested reader to see Stark and Hoey \cite{stark2021ethics}, who discuss how different theoretical conceptualizations of emotion should shape the ethical design of AI systems.

The vast majority of affectively-aware AI is implicitly built upon a basic emotion theory, which assumes that emotions exist as distinct natural categories and can be recognized solely from behaviour. Such AI are usually trained using supervised learning \cite{zeng2009survey} to perform single-example classification of stimuli from single modalities, taken out-of-context (e.g., a still photo), into one of several pre-defined categories. By contrast, real-life emotion detection done by humans is multimodal, evolves over time, and has to be done in context \cite{ong2019modeling}. Although there have been more academic research in recent years on complex multimodal emotion recognition systems \cite{dmello2015review, poria2017review}, commercially-available software are still only unimodal, with the most mature being recognition of (still) facial expressions, such as Affectiva's AffDex, Microsoft's Azure API, and Amazon's Rekognition. Furthermore, because of the cost associated with collecting datasets with a larger variety of emotions, most academic research as well as commerical offerings are trained to only recognize a small number of emotions (e.g., 6), which is hardly representative of real-life.

In a recent comprehensive analysis, \cite{barrett2019emotional} argued that current technology excels at detecting facial \emph{movements}, but the mapping from facial movements to the underlying emotions is not one-to-one but many-to-many\footnote{This is likely the case for \textit{any} single modality, not just facial expressions, which are the most well-studied.}. Because emotions, which are themselves directly unobservable, produce many types of observable behavior, integrating multimodal behavior can help to triangulate what someone is feeling \cite{ong2015affective, anzellotti2019leveraging}. But in many cases, even that is not enough to achieve true emotion understanding. One needs to understand the context \cite{barrett2011context}---is this person watching sports or attending a performance review?---incorporate external world and cultural knowledge, as well as infer people's mental states, their goals, expectations, and even past histories, in order to take their perspective and truly reason about how they feel \cite{ong2019computational, saxe2017formalizing}. Current AI do not take these into account, and so may even be \emph{fundamentally limited} in this endeavour of reading internal states solely from external behavior; we may require substantive changes to AI research paradigms \cite{ong2021applying} in order to achieve AI with a human-centric emotion understanding. 
%
If the entities who develop or deploy such systems assume that the readout of such AI systems are veridical ``emotions" (rather than noisy inferences), without adequately considering context-specificity, social/cultural influences, or even inter- and intra-individual variation, they could end up reading too much into facial movements\footnote{We note also that there are other challenges to the validity of simply ``reading emotions from expressions", namely that people can deliberately alter their facial expressions away from their true emotional states for a variety of reasons such as: adhering to social and cultural display norms, as part of their job (emotional labor), as a strategy in negotiations.}, and make impactful decisions based on sorely incomplete information. 

While there is no doubt that our technology will continue to rapidly improve, we feel that at the present moment, affect recognition technology does not yet deliver what many people believe it to. Consumers of such technology need to treat such AI as a(n incomplete) statistical model, and not a magic crystal ball that perfectly discerns people's hidden emotional lives.




%
%
%
%

\section{An Ethical Framework}

In this section, we outline our proposed ethical framework. 
To the best of our knowledge, this is the first framework that distinguishes the ethical responsibilities of those that \textbf{develop} the AI from those that deploy or \textbf{operate} the AI. 

\begin{figure}[tb]
\begin{center}
\includegraphics[width=.75\columnwidth]{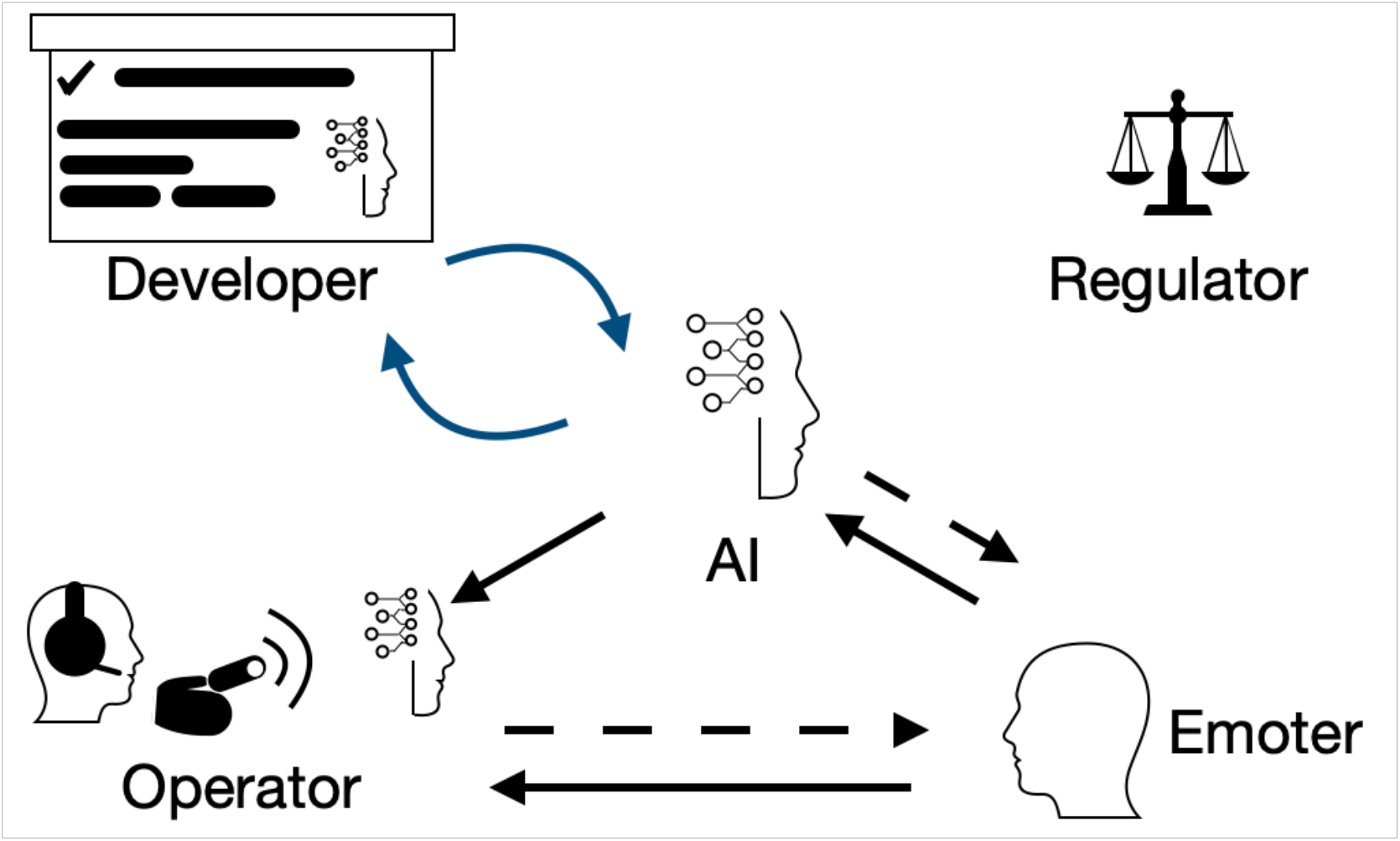}
\end{center}
\caption{Illustration of the four stakeholders involved in an interaction with an affectively-aware AI. Solid arrows indicate flow of information. The Developer trains and produces an AI and delivers it to an Operator. The Operator then deploys the AI to collect and process data from the Emoter. 
The Operator may also collect other data directly from the Emoter. The dashed lines from AI to Emoter, or Operator to Emoter, indicate that, in general, the Emoter may not be aware of the AI or the Operator. The Regulator is also a stakeholder, but does not participate directly in interactions.} 
\label{fig:interactionDiagram}
\end{figure}

First, we identify the four stakeholders in an interaction involving an affectively-aware AI. There is (at least) one individual whose emotions are being read by the AI; we shall call them \textbf{Emoters}. The \textbf{AI} itself could be embodied, as in a social robot, or disembodied, as in the AI powering a system of surveillance cameras. There are the \textbf{Operators}, who are the entities that deploy the AI, and to whom the AI reports the output of any emotion analysis to. The Operator then makes decisions based on the AI's analysis, or they could delegate any subsequent decision-making to other AI systems. Ultimately, because they are the experts in the domain the AI will be deployed in, they hold ethical responsibility for the proper \emph{deployment} of the AI, the \emph{security and collection of the data}, as well as for \emph{decisions resulting from the AI's analysis}. 

The \textbf{Developers} of the AI had created and delivered the AI to the Operators, but may not be involved in every interaction. Because they are the ones designing, training, and maintaining the AI, and are often the only ones that can modify the AI, the Developers thus hold ethical responsibility over the \emph{design and validity of the AI}. Finally, \textbf{Regulators} (i.e., lawmakers) are an important class of stakeholders, who may not be involved in any interaction, but have responsibility for advocating for the welfare of their citizens. We note that this general framework covers special cases where the Developer and Operator can be the same entity, as in the case of the deployment of an in-house developed AI, or where the Regulator is also the Operator, as in the case of surveillance for public safety.

The emerging global consensus, from a recent analysis of global ethics statements \cite{jobin2019global}, is that AI systems should empower people, maximizing well-being and minimizing harm, while treating people fairly with transparency, and respecting people's rights to privacy, autonomy and freedom. 
Although it is obvious that the ethical responsibilities of ensuring this for affectively-aware AI lie with the Developer and Operator, the exact division of labor is less clear-cut. This is undesirable, as it would lead to either party absolving themselves of responsibility, and finger-pointing when an incident happens. Our analysis helps clarify the responsibilities of each party to minimize absolution of responsibility, and will help Regulators to determine appropriately scoped and targeted regulation. In addition, this stakeholder-focused analysis complements earlier academic work that analyzed applications \cite{cowie2012good, mcstay2018emotional} and the fit between emotion theory and AI models \cite{stark2021ethics}.

Our framework rests on two broad ethical ``Pillars", to convey the idea that both are necessary to support ethical application of affectively-aware AI. The first, \textbf{Provable Beneficence}, concerns proving that the AI will benefit the Emoters, which rests on a necessary pre-requisite that the AI is effective at the function it is supposed to serve, and the primary responsibility of upholding this pillar should lie with the Developer. The second, \textbf{Responsible Stewardship}, concerns the actual deployment of the AI, which also includes storing and using the data responsibly; the primary responsibility for upholding this pillar lies with the Operator.

\subsection{Provable Beneficence}

The first ethical principle in our framework is beneficence: the benefit to the Emoter must outweigh the costs to the Emoter, and any such costs must be minimized. 
When dealing with AI systems, we can go one step further and demand \textbf{provable beneficence}; that is, steps must be taken to guarantee, to the best of the Developer's ability, that the AI is beneficial and does no harm. 
Any application of AI should require an analysis of the potential benefits, which depends on specific use-cases (and, in fact, depends also on the Operator---we return to this later). However, a necessary pre-requisite for beneficence is that the AI's predictions must be \emph{credible}; it must agree with reality and must do what it says it does---without which the AI cannot be said to benefit (and not harm) people even if done with the best of intentions.
Thus, provable beneficence entails the following sub-principles: (i) scientific validity, (ii) bias minimization, (iii) generalizability, and (iv) AI transparency and accountability. The responsibility of upholding these sub-principles rests with the Developer.

\subsubsection{Scientific Validity}

In order for an AI to provably benefit the Emoter, its models of emotion must be scientifically valid. Validity refers to the degree to which the AI's measurements or inferences reflects the underlying emotional phenomena it purports to measure. While this may seem obvious, the standards by which validity is assessed differ, even amongst various academic fields. That is why academics hold as the highest gold-standard scientific peer-review done by experts with relevant scientific expertise and who can make proper, contextualized evaluations. 
However, developers do not all reside in academia, and some may not rely on peer-review as a validation strategy, due to concerns about intellectual property and commercial competition. Society must, however, insist on some independent process of determining scientific validity as a pre-requisite for provable beneficence. Some examples include setting up an internal peer-review system, subjecting the AI design principles to independent review or audit by an external board or auditor, accreditation, or testing using randomized controlled trials or on out-of-sample data. 
%

\; \underline{Expression or Emotion?} Scientific validity includes being sensitive to the vast intra-personal, inter-personal, and inter-cultural differences in emotions and emotion expressions \cite{mesquita1992cultural}.  
Additionally, Developers have to acknowledge the importance of recognizing emotions in the context in which they arise, rather than simply classifying stimuli taken out of context.
Many commercially available technology today are unimodal systems that recognize facial expressions from isolated faces. But the mapping from expressions to emotions is complex and a many-to-many mapping
\cite{barrett2019emotional}; without additional modalities or contextual information, unimodal systems are limited in the accuracy that they can achieve.
Developers need to work towards building more comprehensive multimodal, context-aware models, in order to improve the validity of their models.





\begin{figure}[tb]
\begin{center}
\includegraphics[width=.95\columnwidth]{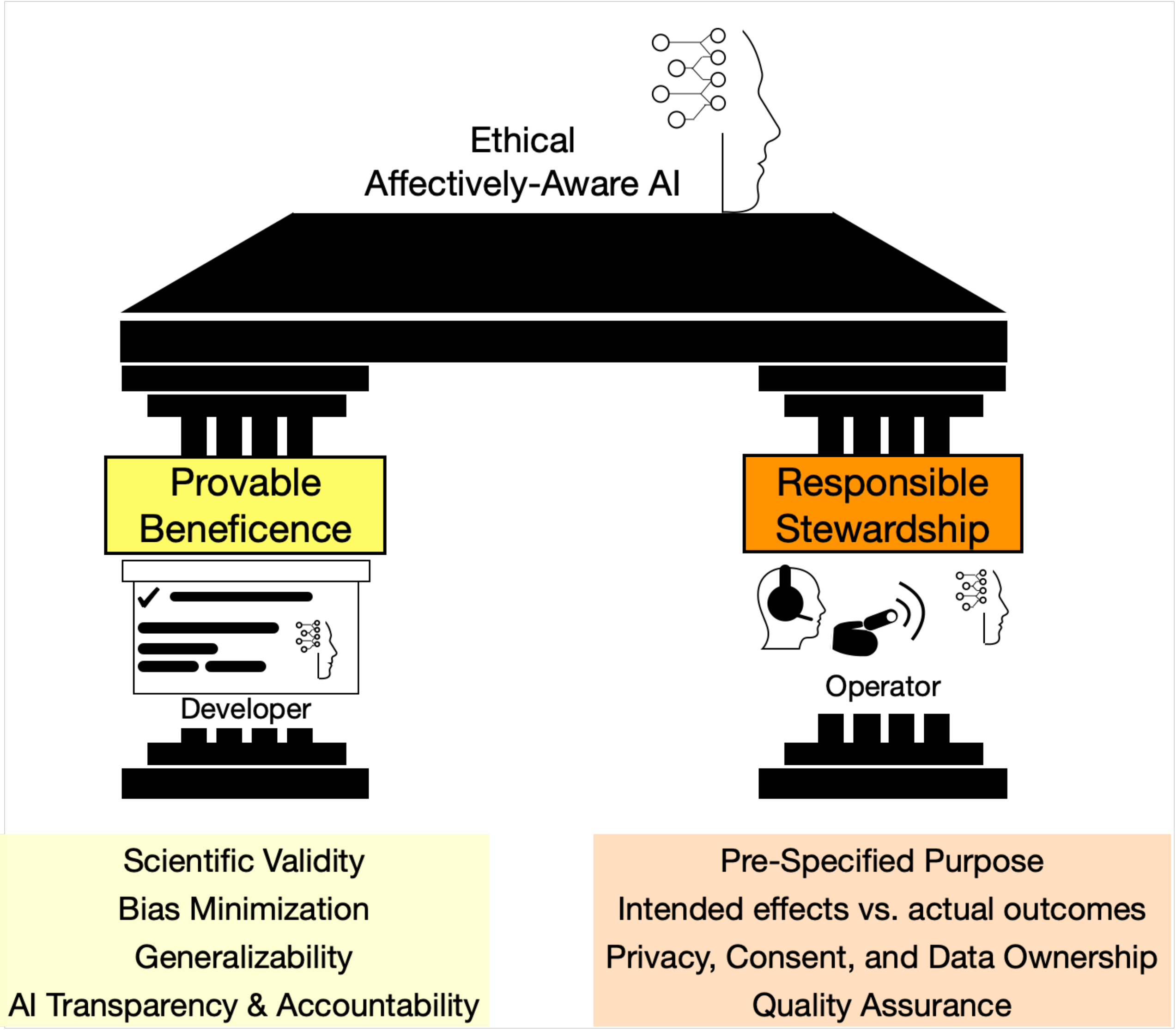}
\end{center}
\caption{The Pillars of Provable Beneficence (upheld by the Developer) and Responsible Stewardship (upheld by the Operator), alongside their sub-principles. These two pillars together support the ethical deployment of affectively-aware AI.} 
\label{fig:pillarDiagram}
\end{figure}

\subsubsection{Bias Minimization}

Developers have to ensure that their AI results in fair treatment of people. To do so, the Developer has to take steps to minimize bias, and to ensure that their data is representative of various groups of people. Machine learning models trained on a finite set of data may learn biases inherent in that data, and thus propagate such bias forward \cite{propublica2016machine}. For example, \cite{rhue2018racial} showed that judgments made by two commercially-available emotion recognition software are biased on race: African American sportsmen are consistently rated as displaying more negative emotions than White Caucasian sportsmen, even when controlling for the intensity of their smiles. These biases could then affect AI-made decisions about, for example, which job applicant to hire, unfairly penalizing certain groups.

Another source of bias is human coders. AI research today involves collecting and labelling large datasets, often through scalable methods like crowd-sourcing on platforms like Amazon Mechanical Turk or Prolific. Relying on untrained coders could lead to data quality problems such as: lack of calibration of ratings and scale-usage, non-standardized understanding of instructions or constructs, cultural differences in emotion concepts \cite{mesquita1992cultural}, language barriers, or distracted and unmotivated coders. Indeed, many researchers acknowledge these issues, and compensate by collecting \emph{more} ratings to average out the noise. This may work for certain idiosyncratic human biases, but if there were systematic biases, such as cultural biases against other races or specific emotions, then crowd-sourcing could further entrench these biases in the data. 

\subsubsection{Generalizability}

A third issue is whether AI models can accurately generalize to new, out-of-sample data. 
%
Developers are incentivized to maximize their model's performance, 
but must also be willing to accept more variance in their data, in order to more accurately capture the vast heterogeneity of human emotional experience and expression. This is especially true for representative data from minorities or vulnerable populations. This poses a dilemma, as collecting higher-variance data will result in a short-term drop in performance metrics like classification accuracy. Developers can justify this decision as ultimately improving their AI in the long-term.

\; \underline{How many emotions?} Existing datasets tend to be limited in their coverage of emotions: most datasets contain a handful of six to eight emotions. Thus, AI models trained on these datasets will be severely handicapped, as they will not know how to recognize other emotions, leading to potential misclassifications.
Recent datasets have sought to expand the number of emotion classes (e.g., 32 classes \cite{rashkin2019towards}), but it is difficult to provide a ``universal" answer of how many categories is enough. The Developer should examine each application in consultation with the domain-expert Operator.


\; \underline{Transfer Learning:} Consider the case where the Operator wants to deploy the AI in a \emph{vastly different setting} or population than the Developer's training data, and which the Developer does not have access to. For example, the Developer's AI is trained mostly on White Caucasian faces in Europe or the US, but the Operator wants to deploy it in a predominantly Asian context (in Asia). Or the Developer has trained their AI on adult faces, but the Operator wants to use it in a specific context with elderly people. 
In such cases, although the responsibility of operation lies with the Operator, the Operator has no access to the inner workings of the AI. The ``know-how" and the ability to test and modify the AI rests with the Developer, but the Developer has no incentive to spend resources to verify their model on a new population. Who then, bears the burden of responsibility for the generalizability to an entirely new population?

We argue that the Operator bears the (ethical and in the future, legal) responsibility of the actual AI's deployment and the decisions made from them. But because they cannot easily re-train the AI or verify the generalizability of the AI on the target population, they must enlist the Developer's help to ensure generalizability, by providing the Developer with data to re-train and to evaluate the AI. 
Thus, the Operator must, as part of contract negotiations, demand proof from the Developer that the predictions of the AI are valid in the Operator's target domain. 
The Developer similarly must work with the Operator to ensure their AI is accurate on this new domain.

\subsubsection{AI Transparency and Accountability}

Unlike other industries like automobiles and aviation, there is no standard regulatory framework for AI. While airplanes and cars have to undergo vigorous safety inspections, there is no similar quality-assurance process for AI that could make impactful decisions. AI Developers should be required to disclose how their technologies were developed, what types of data were they trained on, and what their limitations are. Recent proposals include releasing Model Cards \cite{mitchell2019model} to detail model performance characteristics, including the intended use-cases and contexts, as well as Datasheets for Datasets \cite{gebru2018datasheets} to provide details on the data that the models were trained on (e.g., how were the data collected; what are the demographics of the people involved?). These accompanying reports could be provided as supporting evidence to demonstrate the ``efficacy" of the product, in this case, the AI model. Regulators could verify such claims by requiring regular audits of AI technology.

\; \underline{No snake oil, please! Regulating AI Advertising:} AI Transparency includes being accurate and honest in advertising. The Operator (and Emoter) may not understand the limitations behind affectively-aware AI, and may believe overblown claims by companies about emotion recognition AI \cite{chen2020emotion}. Advertisements for certain classes of regulated products such as medical and financial products must contain tempered claims and disclaimers about potential risks (to health and to financial investments, respectively). Unfortunately AI advertising is not subjected to oversight, and so Developers can advertise unrealistic capabilities that are likely not backed up by evidence. Part of the solution may involve regulation, and one approach could be to study advertising regulation models like in medicine, and mandate that AI advertising similarly include a discussion of limitations, risks, and known weaknesses of the models. Specifically for affectively-aware AI, Regulators should mandate that advertising include a discussion of the demographics of the Emoters in the data that the models were trained on (i.e., information in model cards \cite{mitchell2019model} and datasheets \cite{gebru2018datasheets}), as well as a discussion of context (e.g., what were Emoters in the dataset doing? What situations were they experiencing?).
Another possible solution is self-regulation, where Developers temper their own advertising to potential customers. To justify the short-term cost of more realistic (and less appealing) advertising, Developers should consider that any mismatch between Operator expectations and reality would result in lower consumer confidence in the long-term. 

\subsection{Responsible Stewardship}

Next we turn to the pillar of \textbf{responsible stewardship}. The Operator will be using the AI to collect sensitive, personal data about individual Emoters, and will be making decisions based on the results of analysis on such data. The Operator thus becomes the steward of that data, and has an ethical responsibility to the Emoters to ensure proper use and care of their data. This pillar entails the following four sub-principles: (i) adhering to a pre-specified purpose, (ii) studying whether the intended effects differ from actual outcomes, (iii) being judicious about privacy, consent, and data ownership, and (iv) maintaining quality assurance.

\subsubsection{Pre-specified purpose}

We start with the least familiar idea, which may have the most impact: that of adherence to a pre-specified purpose \cite{microsoft2018responsible}. 
Affectively-aware applications have to be defined with a pre-specified purpose, and subject to Operators’ internal oversight. This will prevent ``mission creep", whereby the same data will gradually be used for different purposes that they may not actually be valid for\footnote{Recent examples criticized as mission creep include deploying counter-terrorism tools for domestic Covid-19 contact tracing \cite{sachs2020technosurveillance} or using contact-tracing data for criminal investigations \cite{tarabay2021countries}.}.

As an example, a bank may start a project to collect emotional information to ``to better understand our customers", which is an underspecified objective. They may initially use emotion expressed by customers during bank visits to improve customer service. But with initial success and without proper oversight over possible uses of the data, they may one day try to use that data to predict credit-worthiness, which may not be a (scientifically) valid use-case. Furthermore, it would be ethically questionable if customers were initially informed of and gave consent to the purpose of improving customer service, but the data was later used to serve other purposes.

In order to properly safeguard Emoters, Operators have to focus on the specific application of such AI, and the specific benefits accrued. This has to be spelled out clearly in their strategy. Internal oversight should ensure that data collected about Emoters are minimal and relevant to the specified purpose (for example, does zip code data need to be tied to emotional expressions to improve customer satisfaction?).


\subsubsection{Distinguish Intended Effects and Actual Outcomes}

Also related to the principle of beneficence, Operators have the responsibility to ensure that the actual outcomes of AI deployment match the intended effects. This entails protocols to continually measure outcomes of interest to ensure that they match the intended effects, and that there are no unintended negative side-effects. This may seem obvious, but it is also tempting from a Operator's point of view to just ``trust" that the AI is working, as there could be a substantial cost involved in monitoring these outcomes (e.g., surveying customers).

This is particularly important if there are potentially vulnerable populations that could be at risk for disparate impact, even if done with the best of intentions. 
For example, if a school decides to implement ``engagement detection" AI tools to improve the quality of education offered, individual teachers may decide to penalize students if they were not paying ``sufficient" attention, perhaps by singling them out in front of the class or giving lower participation grades. Would this unfairly penalize students with attention deficit problems, or who may be going through other difficult personal/family issues? This could be an unintentional side-effect that has a negative impact on students. Thus, Operators have to be aware of how the AI is actually being used on-the-ground and monitor the actual outcomes of their decisions.



\subsubsection{Privacy, Consent, and Data Ownership}
The third set of considerations relates to the interrelated issues of privacy, consent, and data ownership. Given that some emotional information are constantly being ``broadcasted" (much like one’s facial identity) \cite{sedenberg2017smile}, AI Operators need to establish a reasonable standard of privacy for the collection and use of Emoters' facial expressions and other emotional information, while maintaining Emoters' autonomy over their participation in such interactions and data generated from such interactions.

\; \underline{A Culture of Data Consent:} Emoters should have a reasonable expectation of privacy, and should consent to any data collection. In a public space, where it may not be feasible to get individual consent, there should be signs prominently displayed that inform Emoters of the deployment of emotion recognition AI technology. 
This is true even if an Emoter may be reasonably aware that their emotions may be ``read", such as when they are interacting with an embodied AI like a service robot with visible cameras and the ability to display its own expressions. 
The need for highlighting data collection is more pressing when Emoters are reasonably not expecting to have their emotions be ``read", which could happen with disembodied AI, like an AI taking in information from a collection of security cameras. 
In a private setting (in a car cabin, at home, using one's personal device), presumably the Emoter is actively using and interacting with the AI, but the Emoters may not be aware of their sharing emotional information, depending on their expectations of the interaction. In these cases, the Operator should seek explicit consent to collect and use Emoters' emotional expressions.

The exact use of such data should be described clearly to Emoters, and any changes or updates to the software should be reflected explicitly to Emoters. Emoters sould also have the right to opt-out of any data collection, and to request any previously-collected data to be destroyed. Operators should also specify if any data processing is done ``on the edge" (that is, on the device itself, such that the data does not leave the device), or if the data is sent to some external cloud or server.



%

%


\subsubsection{Quality Assurance}

The fourth sub-principle of the Responsible Stewardship pillar is ensuring quality. In a parallel to Developer's efforts to maintain quality, in the context of validity and AI transparency, Operators also need to ensure proper operation of the AI.
This entails ensuring the personnel directly interfacing with the AI have undergone adequate training to use the AI, to correctly interpret the output of the AI, and to troubleshoot any possible errors. 

This extends to safeguarding sensitive Emoter data. Operators should implement proper data control procedures, such as strict data access policies, and implement (cyber)security measures to minimize data leaks.



\section{Discussion}

These principles offer a start in thinking through some of the ethical issues at stake when developing and deploying affective-aware AI. We propose that Developers and Operators should, using this framework, recognize the ethical responsibilities placed upon them due to their respective roles as the AI creator or AI user, and accordingly enact strategy and policy. 

There are, of course, difficult questions when one gets into the details. For example, consider the AI candidate assessment tool discussed in the introduction \cite{fernandez2020ai}. Here, the Operator's utility is not aligned with the Emoters', as the Emoter may feel that their emotional information could be used ``against" them by the Operator and AI. In such cases, how should the risks to the Emoter be weighed against the benefit to the Operator?

Or consider the case for emotion recognition for public safety surveillance, where the benefit is to ``Society" as a whole \cite{standaert2021smile}. Here, the Operator and the Regulator are the same entity; in the former role they may seek more data and less restrictions on their use, while in the latter role they have to weigh the risks to Emoter privacy. Such a balance has to be resolved via a society-wide discussion of the trade-offs that citizens are willing to tolerate in the name of public safety; The answer will differ for each society, and will evolve over time.






\subsection{Recommendations}

For \textbf{Developers}, we recommend developing a practice of getting third-party scientific review for developed technology. Peer review by scientific journals and conferences are ideal, but minimally we recommend an external, rotating board of scientific experts that could include lawyers and ethicists who can review the technology as well as use-cases and advertising. Companies can also put into place policies for formal ethical impact assessments (e.g., \cite{satori2017ethics}).

We also recommend appointing quality-assurance engineers whose job it is to critically test the technology by playing devil's advocate: actively challenging the design of the AI, looking for bias against certain populations, or testing generalizability on different datasets and domains\footnote{This is similar in spirit to `red teams' in cybersecurity which attempt full-scale attacks to test an organization's defenses, but our suggestion differs as the quality-assurance engineer should be a part of the design and engineering process and raise challenges throughout, not just on the final product.}. Designating such an appointment---and designing their incentives and reporting structures appropriately---will, in the long-term, result in more robust and ethical AI. 

For junior developers who work directly with the models and data, we recommend adopting recently-proposed best practices in AI, such as producing detailed Model Cards \cite{mitchell2019model} for their AI models that discuss their model performance specifications, and Datasheets \cite{gebru2018datasheets} that describe the characteristics of the datasets that developers might collect to train their models. Developers could also conduct internal audits, especially on specific demographic groups \cite{bryant2019comparative}. These activities could help junior developers to clearly and accurately convey the AI capabilities to senior executives (and consequently, to external parties like Operators, Emoters, and Regulators).

For \textbf{Operators}, we recommend being involved in discussions of the validity of the AI. During negotiations, challenge the Developer to show proof of validity on the desired use-case, and generalizability on data in the target domain. Operators should also designate (regular) internal oversight for the purpose of data use, and how data is collected and stored. 

We recommend appointing ``consumer advocates" within the organization whose job it is to take the perspective of the Emoter, and challenge project managers on the necessity and use of such data: ``Did the Emoters consent to this new use?"; ``How would you [the manager] feel if you were the Emoter?" These conversations may be difficult to have (and such positions have to be carefully designed so as not to simply stymie activity or add more red tape), but in the long-term they benefit both the Emoter and the Operator.

For \textbf{Regulators}, we recommend appointing experts, whether in-house or from academia, that can be consulted on such AI technology. Other domains (e.g., medicine, finance, energy) are very well-regulated, but AI technology today does not face such similar regulation. Because AI development is so fast-paced, there may also be confusion about current AI capabilities. Thus, regulation needs to be as fast-moving, and quickly adapt to current AI trends, which may suggest adopting a more agile and responsive model of regulation.  

Regulators could also set up an audit program that helps to verify the accuracy of affectively-aware AI. For example, the US National Institutes of Standards and Technology has a long-running Face Recognition Vendor Test (FRVT) program which evaluates vendor-submitted models for facial recognition accuracy under a wide array of conditions such as across demographic groups \cite{grother2019face}. This could serve as a possible model for similarly auditing emotion recognition technology.

Regulators should also consider efforts on the advertising of AI, especially about accurate representation of AI capabilities and limitations. And finally, we recommend that special attention be paid to applications where the Emoter is not in a position of being able to opt-out (e.g., AI-assisted hiring, employee and student monitoring, public safety surveillance).

\section{Conclusion and A Call to Action}

In summary, although there have been many recent conference panels and discussions on the ethics of affectively-aware AI, there has not been much progress by affective computing researchers towards providing a formal, guiding framework. 
In this paper, we propose an ethical framework on which researchers, engineers, industry and regulators, can refer to when evaluating AI applications and deployment. 
Our novel multi-stakeholder analysis separates the burden of responsibilities of AI Developers from the Operators that deploy such AI, and begins to clarify issues for further action by the relevant entities. 

AI Ethics is a habit that the stakeholders, from AI Developers to Operators, from junior engineers to C-suite executives, have to inculcate into their everyday decision-making. We hope that the issues raised will start conversations in individual organizations, and the recommendations will provide concrete, actionable items to work on. We will not achieve ethical (affectively-aware) AI overnight, but it is a shared responsibility that we have to collectively strive for.

\section*{Acknowledgments}

I am grateful to Patricia Chen, Sydney Levine, and three anonymous reviewers for discussions which helped strengthen the paper.


\bibliographystyle{IEEEtran}
\bibliography{biblio}

\begin{thebibliography}{10}
\providecommand{\url}[1]{#1}
\csname url@samestyle\endcsname
\providecommand{\newblock}{\relax}
\providecommand{\bibinfo}[2]{#2}
\providecommand{\BIBentrySTDinterwordspacing}{\spaceskip=0pt\relax}
\providecommand{\BIBentryALTinterwordstretchfactor}{4}
\providecommand{\BIBentryALTinterwordspacing}{\spaceskip=\fontdimen2\font plus
\BIBentryALTinterwordstretchfactor\fontdimen3\font minus
  \fontdimen4\font\relax}
\providecommand{\BIBforeignlanguage}[2]{{%
\expandafter\ifx\csname l@#1\endcsname\relax
\typeout{** WARNING: IEEEtran.bst: No hyphenation pattern has been}%
\typeout{** loaded for the language `#1'. Using the pattern for}%
\typeout{** the default language instead.}%
\else
\language=\csname l@#1\endcsname
\fi
#2}}
\providecommand{\BIBdecl}{\relax}
\BIBdecl

\bibitem{cha2020smile}
S.~Cha, ``{`Smile with your eyes': How to beat South Korea's AI hiring bots and
  land a job},'' \emph{Reuters}, 2020.

\bibitem{metz2020there}
R.~Metz, ``{There's a new obstacle to landing a job after college: Getting
  approved by AI},'' \emph{CNN Business}, 2020.

\bibitem{fernandez2020ai}
C.~Fern{\'a}ndez-Mart{\'\i}nez and A.~Fern{\'a}ndez, ``{AI} and recruiting
  software: {E}thical and legal implications,'' \emph{Paladyn, Journal of
  Behavioral Robotics}, vol.~11, no.~1, pp. 199--216, 2020.

\bibitem{chen2020emotion}
A.~Chen and K.~Hao, ``{Emotion AI researchers say overblown claims give their
  work a bad name.}'' \emph{MIT Technology Review}, 2020.

\bibitem{harwell2019face}
D.~Harwell, ``{A face-scanning algorithm increasingly decides whether you
  deserve the job},'' \emph{Washington Post}, 2019.

\bibitem{ai-video-act-2019}
``Artificial intelligence video interview act of 2019,'' 2019, passed by the
  Illinois General Assembly as Public Act 101-0260 and retrieved from:
  \url{http://www.ilga.gov/legislation/publicacts/fulltext.asp?Name=101-0260}.

\bibitem{dzieza2020emotion}
J.~Dzieza, ``{How hard will the robots make us work?}'' \emph{The Verge}, 2020.

\bibitem{andrejevic2020facial}
M.~Andrejevic and N.~Selwyn, ``Facial recognition technology in schools:
  Critical questions and concerns,'' \emph{Learning, Media and Technology},
  vol.~45, no.~2, pp. 115--128, 2020.

\bibitem{li2019brain}
J.~Li, ``{A "brain-reading" headband for students is too much even for Chinese
  parents},'' \emph{Quartz}, 2019.

\bibitem{barrett2019emotional}
L.~F. Barrett, R.~Adolphs, S.~Marsella, A.~M. Martinez, and S.~D. Pollak,
  ``Emotional expressions reconsidered: {C}hallenges to inferring emotion from
  human facial movements,'' \emph{Psychological Science in the Public
  Interest}, vol.~20, no.~1, pp. 1--68, 2019.

\bibitem{crawford2019ai}
K.~Crawford, R.~Dobbe, T.~Dryer, G.~Fried, B.~Green, E.~Kaziunas, A.~Kak,
  V.~Mathur, E.~McElroy, A.~N. S{\'a}nchez, D.~Raji, J.~L. Rankin,
  R.~Richardson, J.~Schultz, and M.~Whittaker, ``{AI} {N}ow 2019 {R}eport,''
  New York, NY: AI Now Institute, 2019.

\bibitem{cowie2012good}
R.~Cowie, ``The good our field can hope to do, the harm it should avoid,''
  \emph{IEEE Transactions on Affective Computing}, vol.~3, no.~4, pp. 410--423,
  2012.

\bibitem{mcstay2018emotional}
A.~McStay, \emph{{Emotional AI: The Rise of Empathic Media}}.\hskip 1em plus
  0.5em minus 0.4em\relax SAGE, 2018.

\bibitem{cowie2015ethical}
R.~Cowie, ``{Ethical Issues in Affective Computing},'' in \emph{The Oxford
  Handbook of Affective Computing}.\hskip 1em plus 0.5em minus 0.4em\relax
  Oxford University Press, 2015, pp. 334--348.

\bibitem{sedenberg2017smile}
E.~Sedenberg and J.~Chuang, ``Smile for the camera: Privacy and policy
  implications of emotion {AI},'' \emph{arXiv preprint arXiv:1709.00396}, 2017.

\bibitem{stark2021ethics}
L.~Stark and J.~Hoey, ``The ethics of emotion in artificial intelligence
  systems,'' in \emph{Proceedings of the 2021 ACM Conference on Fairness,
  Accountability, and Transparency}, 2021, pp. 782--793.

\bibitem{jobin2019global}
A.~Jobin, M.~Ienca, and E.~Vayena, ``The global landscape of {AI} ethics
  guidelines,'' \emph{Nature Machine Intelligence}, vol.~1, no.~9, pp.
  389--399, 2019.

\bibitem{whittaker2018ai}
M.~Whittaker, K.~Crawford, R.~Dobbe, G.~Fried, E.~Kaziunas, V.~Mathur, S.~M.
  West, R.~Richardson, J.~Schultz, and O.~Schwartz, ``{AI} {N}ow 2018
  {R}eport,'' New York, NY: AI Now Institute, 2018.

\bibitem{ead2016}
``{Ethically Aligned Design: A Vision For Prioritizing Wellbeing With
  Artificial Intelligence And Autonomous Systems, Version 1},'' The IEEE Global
  Initiative for Ethical Considerations in Artificial Intelligence and
  Autonomous Systems., 2016.

\bibitem{inria2014ethique}
``{\'Ethique de la recherche en robotique: Rapport no 1 de la CERNA},''
  Commission de r\'eflexion sur l'\'Ethique de la Recherche en sciences et
  technologies du Num\'erique d'Allistene, 2014.

\bibitem{australia2019ai}
D.~Dawson, E.~Schleiger, J.~Horton, J.~McLaughlin, C.~Robinson, G.~Quezada,
  J.~Scowcroft, and S.~Hajkowicz, ``{Artificial Intelligence: Australia's
  Ethics Framework},'' Data61 CSIRO, Australia, 2019.

\bibitem{france2017how}
``{How Can Humans Keep the Upper Hand? Report on the Ethical Matters Raised by
  AI Algorithms},'' French Data Protection Authority (CNIL), 2017.

\bibitem{france2018for}
``{For a meaningful Artificial Intelligence: Towards a French and European
  Strategy},'' AI 4 Humanity, 2018.

\bibitem{dutch2017human}
``{Human rights in the robot age},'' Rathenau Instituut, 2017.

\bibitem{dutch2018dutch}
``{Dutch Artificial Intelligence Manifesto},'' Special Interest Group on
  Artificial Intelligence, The Netherlands, 2018.

\bibitem{comest2017report}
``{Report of COMEST on Robotics Ethics},'' COMEST/UNESCO, 2017.

\bibitem{campolo2017}
A.~Campolo, M.~Sanfilippo, M.~Whittaker, and K.~Crawford, ``{AI} {N}ow 2017
  {R}eport,'' New York, NY: AI Now Institute, 2017.

\bibitem{mcstay2019guidelines}
A.~McStay and P.~Pavliscak, ``{Emotional Artificial Intelligence: Guidelines
  for Ethical Use.}'' COMEST/UNESCO, 2019.

\bibitem{beavers2017moral}
A.~F. Beavers and J.~P. Slattery, ``On the moral implications and restrictions
  surrounding affective computing,'' in \emph{Emotions and Affect in Human
  Factors and Human-Computer Interaction}.\hskip 1em plus 0.5em minus
  0.4em\relax Elsevier, 2017, pp. 143--161.

\bibitem{gross2011emotion}
J.~J. Gross and L.~Feldman~Barrett, ``Emotion generation and emotion
  regulation: One or two depends on your point of view,'' \emph{Emotion
  Review}, vol.~3, no.~1, pp. 8--16, 2011.

\bibitem{ekman1999basic}
P.~Ekman, ``Basic emotions,'' in \emph{Handbook of Cognition and Emotion},
  T.~Dalgleish and M.~J. Power, Eds., 1999.

\bibitem{moors2013appraisal}
A.~Moors, P.~C. Ellsworth, K.~R. Scherer, and N.~H. Frijda, ``Appraisal
  theories of emotion: State of the art and future development,'' \emph{Emotion
  Review}, vol.~5, no.~2, pp. 119--124, 2013.

\bibitem{mesquita2010contextualized}
B.~Mesquita, ``A contextualized process,'' in \emph{The mind in context},
  B.~Mesquita, L.~F. Barrett, and E.~R. Smith, Eds.\hskip 1em plus 0.5em minus
  0.4em\relax Guilford Press, 2010.

\bibitem{zeng2009survey}
Z.~Zeng, M.~Pantic, G.~I. Roisman, and T.~S. Huang, ``A survey of affect
  recognition methods: Audio, visual, and spontaneous expressions,'' \emph{IEEE
  Transactions on Pattern Analysis and Machine Intelligence}, vol.~31, no.~1,
  pp. 39--58, 2009.

\bibitem{ong2019modeling}
D.~C. Ong, Z.~Wu, T.~Zhi-Xuan, M.~Reddan, I.~Kahhale, A.~Mattek, and J.~Zaki,
  ``Modeling emotion in complex stories: the {Stanford Emotional Narratives
  Dataset},'' \emph{IEEE Transactions on Affective Computing}, 2019.

\bibitem{dmello2015review}
S.~K. D'mello and J.~Kory, ``A review and meta-analysis of multimodal affect
  detection systems,'' \emph{ACM Computing Surveys (CSUR)}, vol.~47, no.~3, pp.
  1--36, 2015.

\bibitem{poria2017review}
S.~Poria, E.~Cambria, R.~Bajpai, and A.~Hussain, ``A review of affective
  computing: From unimodal analysis to multimodal fusion,'' \emph{Information
  Fusion}, vol.~37, pp. 98--125, 2017.

\bibitem{ong2015affective}
D.~C. Ong, J.~Zaki, and N.~D. Goodman, ``Affective cognition: Exploring lay
  theories of emotion,'' \emph{Cognition}, vol. 143, pp. 141--162, 2015.

\bibitem{anzellotti2019leveraging}
S.~Anzellotti, S.~D. Houlihan, S.~Liburd~Jr, and R.~Saxe, ``Leveraging facial
  expressions and contextual information to investigate opaque representations
  of emotions.'' \emph{Emotion}, 2019.

\bibitem{barrett2011context}
L.~F. Barrett, B.~Mesquita, and M.~Gendron, ``Context in emotion perception,''
  \emph{Current Directions in Psychological Science}, vol.~20, no.~5, pp.
  286--290, 2011.

\bibitem{ong2019computational}
D.~C. Ong, J.~Zaki, and N.~D. Goodman, ``Computational models of emotion
  inference in theory of mind: A review and roadmap,'' \emph{Topics in
  Cognitive Science}, vol.~11, no.~2, pp. 338--357, 2019.

\bibitem{saxe2017formalizing}
R.~Saxe and S.~D. Houlihan, ``Formalizing emotion concepts within a bayesian
  model of theory of mind,'' \emph{Current Opinion in Psychology}, vol.~17, pp.
  15--21, 2017.

\bibitem{ong2021applying}
D.~C. Ong, H.~Soh, J.~Zaki, and N.~Goodman, ``Applying probabilistic
  programming to affective computing,'' \emph{IEEE Transactions on Affective
  Computing}, vol.~12, pp. 306--317, 2021.

\bibitem{mesquita1992cultural}
B.~Mesquita and N.~H. Frijda, ``{Cultural Variations in Emotions: A Review},''
  \emph{Psychological Bulletin}, vol. 112, no.~2, p. 179, 1992.

\bibitem{propublica2016machine}
J.~Angwin, J.~Larson, S.~Mattu, and L.~Kirchner, ``{Machine Bias},''
  \emph{ProPublica}, 2016.

\bibitem{rhue2018racial}
L.~Rhue, ``Racial influence on automated perceptions of emotions,'' \emph{SSRN
  3281765}, 2018.

\bibitem{rashkin2019towards}
H.~Rashkin, E.~M. Smith, M.~Li, and Y.-L. Boureau, ``{Towards Empathetic
  Open-domain Conversation Models: A New Benchmark and Dataset},'' in
  \emph{Proceedings of the 57th Annual Meeting of the Association for
  Computational Linguistics}, 2019, pp. 5370--5381.

\bibitem{mitchell2019model}
M.~Mitchell, S.~Wu, A.~Zaldivar, P.~Barnes, L.~Vasserman, B.~Hutchinson,
  E.~Spitzer, I.~D. Raji, and T.~Gebru, ``Model cards for model reporting,'' in
  \emph{Proceedings of the Conference on Fairness, Accountability, and
  Transparency (ACM FAccT)}, 2019, pp. 220--229.

\bibitem{gebru2018datasheets}
T.~Gebru, J.~Morgenstern, B.~Vecchione, J.~W. Vaughan, H.~Wallach,
  H.~Daum{\'e}~III, and K.~Crawford, ``Datasheets for datasets,'' \emph{arXiv
  preprint arXiv:1803.09010}, 2018.

\bibitem{microsoft2018responsible}
``{Responsible bots: 10 guidelines for developers of conversational AI},''
  Microsoft, 2018.

\bibitem{sachs2020technosurveillance}
N.~Sachs and K.~Huggard, ``{Technosurveillance mission creep in Israel’s
  COVID-19 response},'' Brookings Institution, 2020.

\bibitem{tarabay2021countries}
J.~Tarabay, ``{Countries vowed to restrict use of COVID-19 data. For one
  government, the temptation was too great},'' \emph{Fortune}, 2021.

\bibitem{standaert2021smile}
M.~Standaert, ``{Smile for the camera: the dark side of China's
  emotion-recognition tech},'' \emph{The Guardian}, 2021.

\bibitem{satori2017ethics}
\BIBentryALTinterwordspacing
``{SATORI: Stakeholders Acting Together On the ethical impact assessment of
  Research and Innovation},'' 2017. [Online]. Available:
  \url{https://satoriproject.eu}
\BIBentrySTDinterwordspacing

\bibitem{bryant2019comparative}
D.~Bryant and A.~Howard, ``A comparative analysis of emotion-detecting {AI}
  systems with respect to algorithm performance and dataset diversity,'' in
  \emph{Proceedings of the 2019 AAAI/ACM Conference on AI, Ethics, and
  Society}, 2019, pp. 377--382.

\bibitem{grother2019face}
P.~Grother, M.~Ngan, and K.~Hanaoka, ``{Face Recognition Vendor Test (FVRT):
  Part 3, Demographic Effects},'' National Institute of Standards and
  Technology, Tech. Rep. NISTIR 8280, 2019.

\end{thebibliography}

\end{document}